\documentclass[nohyperref]{article}

\usepackage{microtype}
\usepackage{graphicx}
\usepackage{subfigure}
\usepackage{listings}
\usepackage{booktabs} 
\usepackage{amsmath, amssymb, amscd, amsthm, amsfonts}
\usepackage{graphicx}
\usepackage{hyperref}
\usepackage{blindtext}

\usepackage{booktabs} 

\usepackage{amsmath,amsfonts,bm}

\def\eqref#1{equation~\ref{#1}}

\def\1{\bm{1}}

\DeclareMathAlphabet{\mathsfit}{\encodingdefault}{\sfdefault}{m}{sl}
\SetMathAlphabet{\mathsfit}{bold}{\encodingdefault}{\sfdefault}{bx}{n}

\newcommand{\R}{\mathbb{R}}

\DeclareMathOperator*{\argmax}{arg\,max}
\DeclareMathOperator*{\argmin}{arg\,min}

\usepackage{hyperref}
\usepackage{url}
\newtheorem{theorem}{Theorem}

\newtheorem{conjecture}[theorem]{Conjecture}
\newtheorem{definition}[theorem]{Definition}

\def\R {{\mathbb{R}}}

\def\N {{\mathbb{N}}}

\usepackage{hyperref}

\usepackage[accepted]{icml2022_TAGML}

\usepackage{amsmath}
\usepackage{amssymb}
\usepackage{mathtools}
\usepackage{amsthm}
\usepackage{ifthen}

\DeclarePairedDelimiter{\prn}{(}{)}

\usepackage[capitalize,noabbrev]{cleveref}

\theoremstyle{plain}

\theoremstyle{remark}

\usepackage[textsize=tiny]{todonotes}

\icmltitlerunning{Nearest Class-Center Simplification through Intermediate Layers}

\begin{document}

\twocolumn[
\icmltitle{Nearest Class-Center Simplification through 
                Intermediate Layers}



\icmlsetsymbol{equal}{*}

\begin{icmlauthorlist}
\icmlauthor{Ido Ben-Shaul}{yyy,comp}
\icmlauthor{Shai Dekel}{yyy}
\end{icmlauthorlist}

\icmlaffiliation{yyy}{Department of Applied Mathematics
Tel Aviv University, Israel}
\icmlaffiliation{comp}{eBay Research}

\icmlcorrespondingauthor{Firstname1 Lastname1}{first1.last1@xxx.edu}
\icmlcorrespondingauthor{Firstname2 Lastname2}{first2.last2@www.uk}

\icmlkeywords{Machine Learning, ICML}

\vskip 0.3in
]



\printAffiliationsAndNotice{\icmlEqualContribution} 

\begin{abstract}
Recent advances in theoretical Deep Learning have introduced geometric properties that occur during training, past the Interpolation Threshold- where the training error reaches zero. We inquire into the phenomena coined \emph{Neural Collapse} in the intermediate layers of the networks, and emphasize the innerworkings of Nearest Class-Center Mismatch inside the deepnet. We further show that these processes occur both in vision and language model architectures. Lastly, we propose a Stochastic Variability-Simplification Loss (SVSL) that encourages better geometrical features in intermediate layers, and improves both train metrics and generalization.
\end{abstract}

\section{Introduction}
Several recent works have investigated the nature of modern Deep Neural Networks (DNNs) past the point of zero training error \cite{Belkin2021FitWF, Nakkiran2020DeepDD, Bartlett2021DeepLA, Power2022GrokkingGB}. The stage at which the training error reaches zero is called the \emph{Interpolation Threshold (IT)}, since at this point, the learned network function interpolates between training samples. This is not to be confused with zero-loss, but simply the point where all training samples are correctly classified. The stage of training beyond the IT is coined the \emph{Terminal Phase of Training (TPT)} in \cite{Papyan2020PrevalenceON}. It was in this paper that the term \emph{Neural Collapse (NC)} was introduced to describe four interconnected geometrical phenomena that describe the network behavior past the TPT. Let us briefly describe the properties of NC that are most relevant for this paper:

\textbf{(NC1) Variability collapse:} As training progresses, the
within-class variation of the activations becomes negligible
as these activations collapse to their class-means.

\textbf{(NC4) Simplification to Nearest Class-Center (NCC):}  For a given deepnet activation, the network classifier  converges to choosing whichever class has the nearest  train class-mean (in standard Euclidean distance).

In this work we delve deeper into the inner workings of NCC Simplification. Several works have proposed that geometrical properties of intermediate layers shape the way deepnets are trained, and largely affect their successes \cite{Separability_and_geometry, AllenZhu2020BackwardFC, BenShaul2021SparsityProbeAT, Liu2020ExplainingDN, LP, Baldock2021DeepLT}. We explore along the lines of \cite{Papyan2020PrevalenceON} to further understand such geometries. 

\subsection{Our contributions}
\begin{figure*}[t]
\begin{center}
   \includegraphics[width=0.95\linewidth]{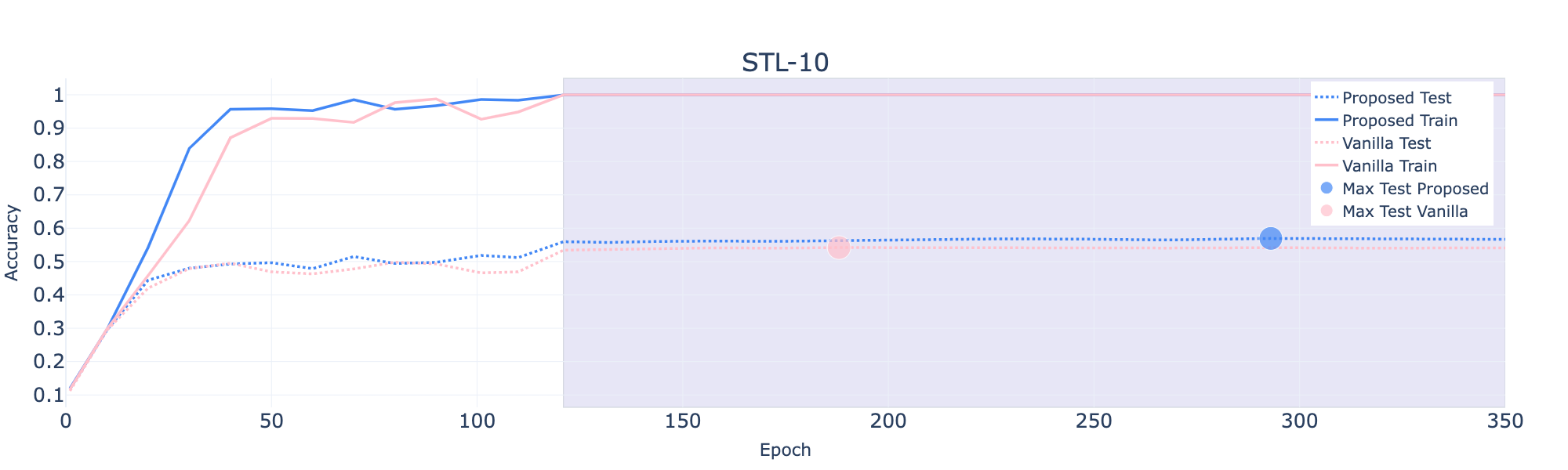}
\end{center}
   \caption{\textbf{STL-10} experiment training procedure. We show the accuracy on both the train and test set, using the different losses: Vanilla Cross-Entropy (\textbf{Pink}) and SVSL (\textbf{Blue}). The SVSL loss outperforms the vanilla in both metrics. Both models improve in terms of test-performance during the TPT, shown as the shaded area. In the TPT, both models have a constant accuracy of 1.}
\label{STL10_training_parts}
\label{fig:long}
\label{fig:onecol}
\end{figure*}

Our contributions can be summarized as follows:

\textbf{NCC Simplification in Intermediate Model Layers:} We show that when looking at the NCC Mismatch of deepnet intermediate layers, before and during TPT, there is a beautiful geometric structure that emerges. Namely:

\begin{enumerate}
\item [(i)] There is a clear ordering between NCC mismatch in intermediate layers. The mismatch is lower as the layers gets deeper.
\item [(ii)] NCC Simplification is not only apparent in the final layer of the network, and may propagate back several layers in the network. 
\end{enumerate}

\textbf{NCC-Simplification is apparent in in Transformer NLP architectures:}
When proposed in \cite{Papyan2020PrevalenceON}, the authors show that Neural Collapse appears in several well-known Image Classification models: VGG~\cite{Simonyan2015VeryDC}, ResNet~\cite{He2016DeepRL}, and DenseNet~\cite{Huang2017DenselyCC} on Image Classification datasets. In this paper, we show that NCC simplification is also apparent in Transformer architectures \cite{Vaswani2017AttentionIA}, and even more surprisingly, in common NLP tasks. The recent surge in Transformer architectures in cross-modal tasks, suggests that there are common behaviors between classic Image architectures, and more recent mechanisms \cite{Radford2021LearningTV, Dai2021CoAtNetMC, Raghu2021DoVT}. 

\textbf{Encouraging Variability Simplification can assist in training and generalization:}
We propose a simple intermediate layer variance collapsing loss which we coin the Stochastic Variability Simplification Loss (SVSL). This loss is shown to improve the performance of a wide-variety of tasks by encouraging NCC simplification during training. We show that this loss is able to improve NCC mismatch between intermediate layers, on both the train and test datasets. The different training stages and their respective metrics can be seen for both losses on the STL-10 dataset in Figure \ref{STL10_training_parts}. The same plot is given for all other datasets in the appendix. In all plots in the paper, the $x$-axis represents the epochs during train. We share the code for reproducing the paper experiments in the supplementary materials. 

\section{Problem Setup and Background}
\subsection{Supervised Classification}
\begin{figure*}[t]
\begin{center}
   \includegraphics[width=1.\linewidth]{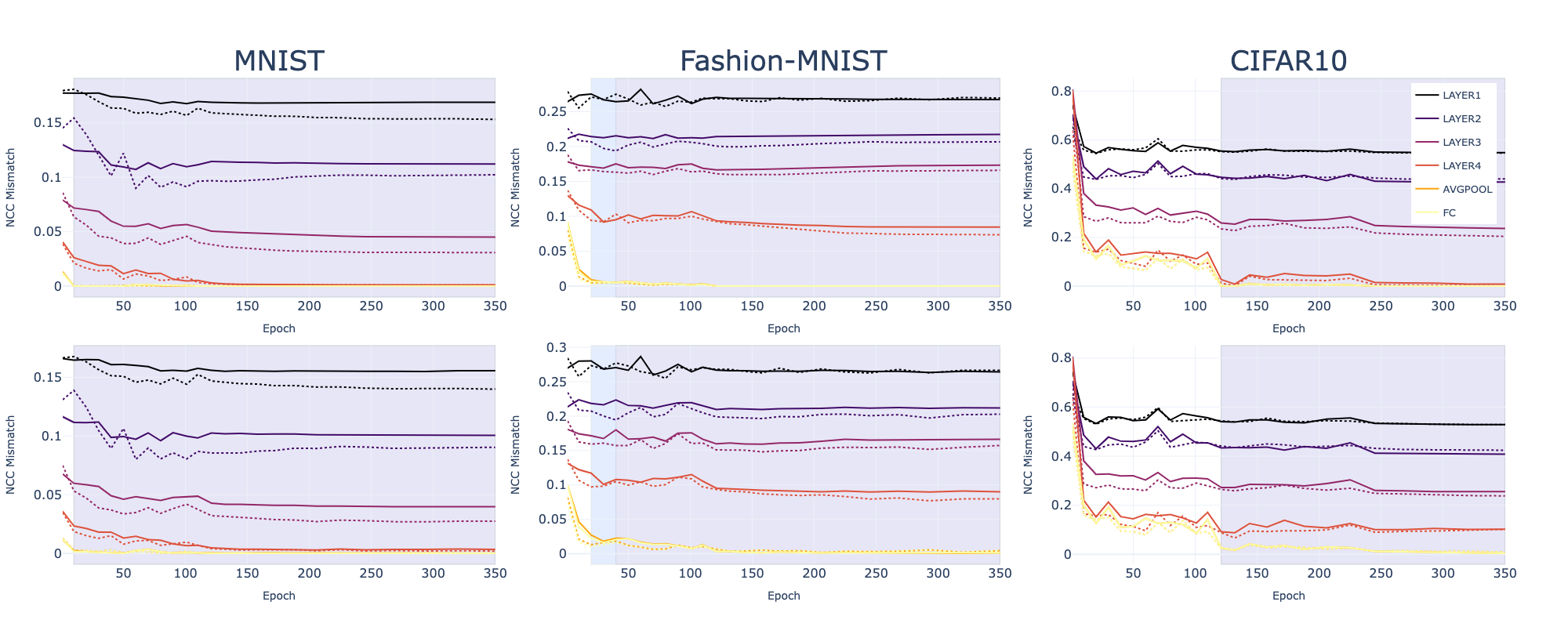}
\end{center}
   \caption{NCC Mismatch for Vision datasets: \textbf{MNIST}, \textbf{Fashion-MNIST} and \textbf{CIFAR10}, using both vanilla (\textbf{solid}) and SVSL (\textbf{dashed}) losses. \textbf{Top:} train NCC Mismatch, \textbf{Bottom:} test NCC Mismatch. The shaded pink background shows the TPT for the vanilla loss experiment, and the blue for the SVSL. The background is shaded purple at epochs when both experiments are in the TPT.}
\label{NCC_MNIST_FMNIST_CIFAR}
\label{fig:long}
\label{fig:onecol}
\end{figure*}
We are interested in the supervised classification setting. In this paper, our experiments include problems in both Image and NLP- Sequence Classification. In \textbf{Image Classification} we are given a training set of $d:={3}\times{W}\times{H}$-dimensional RGB images, from $C$ categories. We wish to train a network to differentiate between different image classes. On the other hand, in \textbf{Text-Sequence Classification} we are given a training set of sequences. We use standard tokenization techniques to transfer discrete sequences to a continuous euclidean space. Since different sequences may be in different lengths, we pad sequences in our experiments to a certain $d:=\mathrm{MAX\_PAD}$ constant. Similar to the Image setting, we wish to find the appropriate class for each text-sequence(of dimension $\R^d$, for $C$ ground-truth classes). 

Let $g$ represent a deepnet, $g:\R^{d}\rightarrow{\R^C}$ where $C$ is the number or classes, and network parameters $\theta$ which are learned through the optimization procedure, or ``training''. For the classification setting, the classification decision of the network for input $x\in{\R}^d$ is defined as $z = \argmax_{1\leq{c'}\leq{C}} {g\left(x\right)}_{c'}$. Let $L=\{l^{\left(1\right)},\dots,l^{\left(k\right)}\}$ represent the set of intermediate layers of network $g$ (formally defined in Section \ref{Optimization}), such that $g:=l^{\left(k\right)}\circ\dots\circ{l^{\left(1\right)}}$. We define $n_j$ as the output dimension of layer $j$, such that $n_0:=d$ and $n_k:=C$. We note $g^{\left(j\right)}\left(x\right)$ as the outputs of the $j^{\mathrm{th}}$ layer of the network for input sample $x$. Using these definitions, the following holds: $g^{\left(j\right)}(x):= l^{\left(j\right)}\circ\dots\circ{l^{\left(1\right)}}\left(x\right)$. Given a function $f$ and a set of indices $\mathcal{I}$ we introduce the streamlined notation $f_i:=f\prn{x_i}$.

\subsection{Representation Learning}
In recent years, the fields of Vision and NLP have both been transformed by representation learning methods in supervised and unsupervised tasks. The main premise is to learn ``representations such that similar samples stay close to each other, while dissimilar ones are far apart'' \cite{weng2021contrastive}. Encouraging clustering in features learned by a deepnet has been a pivotal early idea to improve representations \cite{Xie2016UnsupervisedDE, Tian2017DeepClusterAG}. In this paper we will show that encouraging clustering of intermediate layers can boost performance. We use the normal convention of calling a sample the \emph{Anchor}, a similar example \emph{Positive} and a dissimilar example as \emph{Negative}. In Vision, representation learning can be approached by using clustering assignments as pseudo-labels \cite{Tian2017DeepClusterAG}, invoking similarity between different augmentation of the same sample~\cite{Zbontar2021BarlowTS, Chen2020ASF, Caron2021EmergingPI}, or even using class labels to wisely pick Positive and Negative items~\cite{Khosla2020SupervisedCL}. In Language Modeling(LM), Masking and Next-Sentence prediction(among others) tasks are used to learn semantically robust representations~\cite{Devlin2019BERTPO, Liu2019RoBERTaAR}. We use the notion of representation learning by enforcing low inner-class variability. The methods mentioned only use the penultimate layer of the encoder to penalize the representations. We conjure that better consistency of class representations in intermediate layers forces representations in final layers to have better geometrical features. Our method does not need to sample pairs (positive or negative), and is simple to implement.

\section{Neural Collapse}
We shall now briefly present the particular properties of Neural Collapse that are relevant for this work, as presented in \cite{Papyan2020PrevalenceON, Han2021NeuralCU}.
Let $g$ be a given network and $\{\left(x_i, y_i\right)\}_{i\in{\mathcal{I}_{\mathrm{Train}}}}$, $\{\left(x_i, y_i\right)\}_{i\in{\mathcal{I}_{\mathrm{Test}}}}$ be the train and test set accordingly.
\begin{definition}[Train class-means]\label{train_class_means} We define the train class means for layer $l^{\left(j\right)}$ and class $1\leq{c}\leq{C}$ as 
\[
\mu_{c}^{\left(j\right)} := \mathrm{Avg}_{i\in{\mathcal{I}_{\mathrm{Train}}}, y_i=c}\{g^{\left(j\right)}_i\}.
\]
\end{definition}
\begin{definition}[Train within-class covariance]\label{train_class_means} We define the train within-class covariance for layer $l^{\left(j\right)}$ as 
\[
\Sigma_{W}^{\left(j\right)} := \frac{1}{C}\sum_{c=1}^{C} \mathrm{Avg}_{i\in{\mathcal{I}_{\mathrm{Train}}}, y_i=c}\{(g^{\left(j\right)}_{i}-\mu_{c}^{\left(j\right)})(g^{\left(j\right)}_{i}-\mu_{c}^{\left(j\right)})^\top\}.
\]
\end{definition}
Using these definitions, we can formally define properties (NC1) and (NC4) from  \cite{Papyan2020PrevalenceON}. Let us assume that the network $g$ may be split into two stages: the ``feature engineering'' stage $g^{\left(k-1\right)}$ and the final classifier layer, $l^{\left(k\right)}$ such that $g:=l^{\left(k\right)}\circ{g^{\left(k-1\right)}}$.

\begin{definition}[\textbf{NC1} Variability Collapse]\label{NC1} $\Sigma_{W}^{\left(k-1\right)}\rightarrow{0}$.
\end{definition}

\begin{definition}[\textbf{NC4} Simplification to NCC]\label{NC4}
Let:
\[
\begin{aligned}
{S}&:=\left\{ i\in \mathcal{I}_{\mathrm{Train}} \mid \argmax_{1\leq{c}\leq{C}} {g\left(x_i\right)}_{c} \right. \\&\neq\left.\argmin_{1\leq{c}\leq{C}}{\left\|g^{\left(k-1\right)}\left(x_i\right)-\mu^{\left(k-1\right)}_{c}\right\|_2}\right\},
\end{aligned}
\]
then 
$\left|S\right|\rightarrow{0}$, where $\left|X\right|$ is the number of elements in a finite set $X$. 
\end{definition}
In both Definitions, the $\rightarrow$ is defined as the progress with the optimization procedure. Throughout this paper, we make the assumption that the deepnets are of proper capacity to reach the TPT, or in other words to ``fit'' the data. In the experiments we use large architectures that are highly overparameterized for the given tasks.

\section{Contributions}
\subsection{NCC mismatch in Intermediate Layers}

The results of \cite{Papyan2020PrevalenceON} show a clear behavior in terms of NCC-Simplification in the penultimate layer (see Definition \ref{NC4}). We further investigate the behavior of intermediate layers in terms of NCC mismatch. Let us define:
\begin{definition}[Layer $j$ \textbf{Train} NCC mismatch]
\begin{equation}
\begin{aligned}
\Lambda_{\mathrm{Train}}^{\left(j\right)}&:=\frac{1}{N_{\mathrm{Train}}}\left|\left\{ \argmax_{1\leq{c'}\leq{C}} {g\left({x_i}\right)}_{c'}\right.\right. \\&\neq\left.\left.\argmin_{c'}{\left\|g^{\left(j\right)}\left(x_i\right)-\mu_{c'}^{\left(j\right)}\right\|_2} \mid i\in\mathcal{I}_{\mathrm{Train}}\right\}\right|.
\end{aligned}
\end{equation}
\end{definition}

\begin{definition}[Layer $j$ \textbf{Test} NCC mismatch]
\begin{equation}
\begin{aligned}
\Lambda_{\mathrm{Test}}^{\left(j\right)}&:=\frac{1}{N_{\mathrm{Test}}}\left |\left\{ \argmax_{1\leq{c'}\leq{C}} {g\left({x_i}\right)}_{c'}\right.\right. \\&\neq\left.\left.\argmin_{c'}{\left\|g^{\left(j\right)}\left(x_i\right)-\mu_{c'}^{\left(j\right)}\right\|_2} \mid i\in \mathcal{I}_{\mathrm{Test}} \right\}\right|.
\end{aligned}
\end{equation}
\end{definition}

\underline{\textbf{Our conjectures can now be described as follows}}
\begin{conjecture}[Intermediate Layer ordering using NCC mismatch]\label{NCC_ordering} There is a clear order between both \textbf{train} and \textbf{test} NCC mismatch in intermediate layers. The mismatch is lower as the layers gets deeper. In the TPT, for $1\leq{j}\leq{k}$,
\begin{equation}
\begin{aligned}
\Lambda_{\mathrm{Train}}^{\left(j\right)} \geq \Lambda_{\mathrm{Train}}^{\left(j+1\right)}
\quad\quad\mathrm{and} \quad\quad
\Lambda_{\mathrm{Test}}^{\left(j\right)} \geq \Lambda_{\mathrm{Test}}^{\left(j+1\right)}.
\end{aligned}
\end{equation}
\end{conjecture}

\begin{conjecture}[NCC mismatch improves in TPT]\label{NCC_improves_in_TPT}
At each intermediate layer, both the \textbf{train} and \textbf{test} NCC mismatch improves from the IT to End of Training(EOT).
\begin{equation}
\begin{aligned}
\Lambda_{\mathrm{Train, IT}}^{\left(j\right)} \geq \Lambda_{\mathrm{Train, EOT}}^{\left(j\right)}
\quad\mathrm{and}\quad
\Lambda_{\mathrm{Test, IT}}^{\left(j\right)} \geq \Lambda_{\mathrm{Test, EOT}}^{\left(j\right)}.
\end{aligned}
\end{equation}
\end{conjecture}



\subsection{Decreasing NCC Mismatch using Stochastic Variability-Simplification Loss}\label{loss_function}
In \cite{Papyan2020PrevalenceON}, the properties shown in Definitions \ref{NC1} and \ref{NC4} both act on the penultimate layer of the network. When considering intermediate feature spaces, the property shown in Definition \ref{NC1} amounts to promoting class clustering. Promoting class clustering can push samples further from decision boundaries between classes, and could therefore increase agreement between the nearest class-center and the classifier. We wish to decrease the NCC mismatch during train, and encourage better clustering through the intermediate layers. Our loss function is proposed as follows:

\begin{definition}[Stochastic Train class-means]\label{stochastic_train_class_means} Let $\mathcal{B}:=\left\{\left(x_i, y_i\right)\right\}_{i\in{B}}$, where $\left|B\right|$ is the Batch-Size. We define the stochastic train class means for layer $l^{\left(j\right)}$, batch $\mathcal{B}$, and class $1\leq{c}\leq{C}$ as 
\[
\mu_{c, \mathcal{B}}^{\left(j\right)} := \mathrm{Avg}_{i\in{\mathcal{B}}, y_i=c}\{g^{\left(j\right)}_{i}\}.
\]
\end{definition}

\begin{definition}[Stochastic Variability-Simplification Loss (SVSL)]
Let $g$ be a deepnet and $\hat{y_i}=g(x_i)$ for $\left(x_i, y_i\right), i\in{\mathcal{I}_{\mathrm{Train}}}, y_i=c, 1\leq{c}\leq{C}$. Let $\mathcal{B}$ be the batch such that $i\in{B}$. We also define $\gamma\in\N, 1\leq\gamma<{k}$ and $\alpha\in\R_{+}$ two hyperparameters. The Stochastic Variability-Simplification Loss function is then defined as
\begin{equation}
\begin{aligned}
\mathcal{L}\left(\hat{y_i}, y_i\right) &:= \mathrm{CE}\left(\hat{y_i}, y_i\right) + \eta\sum_{j=\gamma}^{k}{\left\|{g^{\left(j\right)}\left(x_i\right)-\mu_{c, \mathcal{B}}^{\left(j\right)}}\right\|_{2}^{2}}.
\end{aligned}
\end{equation}
where CE is the well-known Cross-Entropy loss and 
\[
\eta = \frac{\alpha}{C\left(k+1-\gamma\right)\left|\{i\in{B} \mid{y}_i = c\}\right|}.
\]

\end{definition}
The normalizing factor $\eta$ serves as a mitigating factor in the case of unbalanced batches. It is possible to define the Variability-Collapse in a non-stochastic fashion, by computing the full class-means at layer for every epoch. An example implementation of the SVSL is given in the appendix.

\underline{\textbf{Using the SVSL, we claim the following behaviors:}}

\begin{conjecture}[SVSL improves NCC mismatch]\label{VSL_Mismatch}
Using the properly defined hyperparameters $\alpha, \gamma$, the Stochastic Variability-Simplification Loss encourages lower \textbf{train} and \textbf{test} NCC mismatch in intermediate layers. In the TPT, for $1\leq{j}\leq{k}$,
\begin{equation}
\begin{aligned}
\Lambda_{\mathrm{Train, Vanilla}}^{\left(j\right)} \geq \Lambda_{\mathrm{Train, SVSL}}^{\left(j\right)}
\quad\mathrm{and} \quad \\
\Lambda_{\mathrm{Test, Vanilla}}^{\left(j\right)} \geq \Lambda_{\mathrm{Test, SVSL}}^{\left(j\right)}.
\end{aligned}
\end{equation}
\end{conjecture}

\begin{conjecture}[SVSL can improve test-performance]\label{VSL_test}
The EOT test metrics are improved for all datasets using the SVSL and proper hyperparameter tuning.
\end{conjecture}
\subsubsection{Motivation for SVSL}\label{motivation}\emph{The Folding Ball Hypothesis} is presented in \cite{10.5555/3203489} as follows: ``Imagine two sheets of colored paper: one red and one blue. Put one on top of the other. Now crumple them together into a small ball. That crumpled paper ball is your input data, and each sheet of paper is a class of data in a classification problem. What a neural network is meant to do is figure out a transformation of the paper ball that would uncrumple it, so as to make the two classes cleanly separable again''. This geometrical notion has been used to try to predict the wellness of such transformations, using their geometrical properties \cite{Separability_and_geometry, BenShaul2021SparsityProbeAT, LP, Montavon2011KernelAO}. When measuring NCC mismatch during TPT, the network has near $0$-training-error. This essentially means that the final feature space(where the inputs to the classifier reside) has near perfect clusters per-class. In \cite{Papyan2020PrevalenceON}, it is empirically shown that for most deepnets, the penultimate layer has a single cluster for each class. Thus, measuring the \textbf{train} NCC mismatch between the $j$th feature-space and the classifier is similar to checking the NCC mismatch with the ground-truth labels. The clustering of feature spaces is an iterative transformation from each layer to the next, where the quality of clustering assists in clustering at the following stage. 

Demanding a low NCC mismatch in early layers of the network may be unsatisfiable, as the input samples (e.g. images) cannot necessarily be well clustered with such low capacity (small number of layers). This is the reason we allow the $\gamma$ hyperparameter to facilitate the earliest layer from which we require the SVSL. Demanding consistency between the NCC and the classifier early in training can interfere with the model learning the proper class predictions, so we leverage between the losses using the $\alpha$ hyperparameter. A different approach can consist of applying the SVSL only during TPT.

\section{Experiment Details}
Our experiments aim to demonstrate Conjectures \ref{NCC_ordering},\ref{NCC_improves_in_TPT},\ref{VSL_Mismatch},\ref{VSL_test} on both Vision and NLP tasks. In Section \ref{datasets} we introduce the datasets that were used. In Section \ref{Architecures} we describe the architectures and Section \ref{Optimization} goes through the training procedures used.


\subsection{Datasets}\label{datasets}
For \textbf{Vision} tasks, we use most of the datatsets used in \cite{Papyan2020PrevalenceON}. Namely: MNIST, FashionMNIST, CIFAR10, CIFAR100, and STL10. Unlike in \cite{Papyan2020PrevalenceON}, we do not balance the datasets explicitly and keep them as they are. We use mean-std train normalization. In order to get intermediate features, we use PyTorch Hooks~\cite{Paszke2019PyTorchAI}. For the \textbf{NLP sequence classification tasks} we use a subset of binary datasets from the GLUE benchmark \cite{Wang2018GLUEAM}. We run our experiments on datasets from all three types of tasks: \emph{Single-Sentence Tasks:} CoLA and SST-2, \emph{Similarity and Paraphrase Tasks:} MRPC, and \emph{Inference Tasks:} RTE. All datasets have $2$ classes. In order to make all sequences of the same length, both for computing NCC mismatch and maintaining same size features, we pad each of the sequences in all datasets to $32$ tokens. Intermediate features are readily given as ``$\mathrm{hidden\_states}$'' in \cite{wolf-etal-2020-transformers}.

\subsection{Architectures}\label{Architecures}
\textbf{Vision}: For the vision architectures we follow the guidelines set in \cite{Papyan2020PrevalenceON}. In this paper we use solely the ResNet~\cite{He2016DeepRL} architectures. ResNet18 is used for MNIST, FashionMNIST, and CIFAR10. For CIFAR100 and STL10 the model chosen is the ResNet50 architecture. The layers for the ResNet architecture that are used in the experiments are $\{\mathrm{Layer1, Layer2, Layer3, Layer4, AvgPool, FC}\}$ as implemented in TorchVision \cite{10.1145/1873951.1874254}. \textbf{NLP Sequence Classification:} For all sequence classification task we use an Uncased pre-trained BERT \cite{Devlin2019BERTPO}. The layers used for this architecture are the hidden states of the BERT-architecture. We include the embedding-layer in the BERT architecture, and use all hidden-states except the final output layer. In total, we have an initial embedding features ($1$) and ($11$) hidden-state layers, for a total of $12$ layers in this architecture. In theory, the penultimate layer can also be used in the optimization process.

\subsection{Optimization Procedure}\label{Optimization}
\underline{\textbf{Vision:}} We use the same optimization scheme as in \cite{Papyan2020PrevalenceON}, using best training hyperparameters as logged, and follow the same training procedure. We train all datasets for 350 Epochs. The Batch-Size for all experiments is $128$. All vision experiments are trained using a SGD optimizer as done in the original paper. All SVSL Hyperparameters used are given in Table \ref{tab:NLP_VSL_HP} in Appendix C. We report the top-1-accuracy on the test datasets. We use a threshold of $0.995$ for determining the Interpolation Threshold.

\underline{\textbf{NLP Sequence Classification:}}
We follow the default hyperparameters as shown in \cite{wolf-etal-2020-transformers} (GLUE finetune example). All experiments are trained using an AdamW~\cite{Loshchilov2019DecoupledWD} optimizer, and the default hyperparameters for $10$ epochs. The Batch-Size for all experiments is $8$, and the tasks are all binary classification. We report test-accuracy for the datasets: RTE, SST-2, and MRPC, and Matthew's-Correlation for the CoLA dataset. \textbf{SVSL Hyperparameters:} The SVSL parameters are found using a simple baysean optimization scheme (AX-BoTorch~\cite{balandat2020botorch}) for  $\alpha\in[5e-8, 5e-5]$ and layers $\gamma\in\{1,\dots,11\}$ on the test set. The purpose of these experiments is to show the ability to improve the network behavior using the SVSL. Possible future research includes adding the hyperparameters to as part of the network weights. The hyperparameters used are recorded in Table \ref{tab:NLP_VSL_HP}. We use a threshold of $0.985$ for determining the IT.

\section{Results}

\subsection{NCC mismatch Behavior in Intermediate Layers}
We wish to demonstrate the Conjectures \ref{NCC_ordering} and \ref{NCC_improves_in_TPT}. The train and test NCC mismatch metrics for the MNIST, Fashion-MNIST, and CIFAR10 datasets are visualized in Figure \ref{NCC_MNIST_FMNIST_CIFAR} (Solid Lines). The same metrics for the sequence classification and all Vision experiments are given in the appendix.

\subsection{Variability-Simplification Loss}
\begin{table*}\centering
\caption{Comparing the Test Metrics of CE-Loss(Vanilla) with Stochastic Variability-Simplification Loss (SVSL) at IT, EOT, and Best-Test-Epoch for both Image and NLP-Sequence Classification Datasets. The metrics are as defined in Section \ref{Optimization} in percents. The Matthew's -Correlation metric for the CoLA dataset is multiplied by a factor of $100$. We also note whether the best Test-Metrics are achieved in TPT, for all datasets and methods.}\vspace{0.1in}
\begin{tabular}{l|ll|ll|llll}
\multicolumn{1}{c|}{} & \multicolumn{2}{c|}{\textbf{IT}} & \multicolumn{2}{c|}{\textbf{EOT}} & \multicolumn{4}{c}{\textbf{Best Test Epoch}}                           \\
Dataset               & Vanilla         & SVSL           & Vanilla      & SVSL               & Vanilla        & \multicolumn{1}{l|}{In TPT} & SVSL           & In TPT \\ \hline
MNIST                 & \textbf{99.37}  & \textbf{99.36} & 99.61        & \textbf{99.69}     & 99.65          & \multicolumn{1}{l|}{Yes}    & \textbf{99.69} & Yes    \\
Fashion MNIST         & 91.78  & \textbf{93.13} & 93.82        & \textbf{93.88}     & 93.93          & \multicolumn{1}{l|}{Yes}    & \textbf{94.03} & Yes    \\
STL10                 & 53.41           & \textbf{55.95} & 54.11        & \textbf{56.65}     & 54.19          & \multicolumn{1}{l|}{Yes}    & \textbf{56.94} & Yes    \\
CIFAR10               & \textbf{80.64}  & 80.56          & 80.96        & \textbf{81.19}     & 80.96          & \multicolumn{1}{l|}{Yes}    & \textbf{81.19} & Yes    \\
CIFAR100              & 52.77           & \textbf{53.28}  & 53.31        & \textbf{54.29}     & 53.79          & \multicolumn{1}{l|}{Yes}    & \textbf{54.29} & Yes    \\ \hline
CoLA                  & 51.59           & \textbf{52.91} & 53.46        & \textbf{55.54}     & 53.95          & \multicolumn{1}{l|}{No}    & \textbf{55.54} & Yes    \\
RTE                   & \textbf{58.84}  & 58.12          & 55.23        & \textbf{59.57}     & \textbf{61.01} & \multicolumn{1}{l|}{No}     & 60.28          & Yes    \\
MRPC                  & 70.83           & \textbf{74.26} & 74.26        & \textbf{75.25}     & 75.00          & \multicolumn{1}{l|}{No}     & \textbf{76.71} & No     \\
SST-2                 & 87.96           & \textbf{88.42}  & 88.42        & \textbf{88.76}     & \textbf{89.22} & \multicolumn{1}{l|}{No}     & \textbf{89.22} & Yes   
\end{tabular}
\label{tab:test_vsl}
\end{table*}
In this section we wish to demonstrate how using the intermediate-layer SVSL can improve training procedure and generalization. In Section \ref{motivation} we describe the underlying logic behind the proposed cost. We advocate that in networks where intermediate NCC mismatch is lower, perform better in the TPT stage. Let us first demonstrate the correctness of Conjecture \ref{VSL_Mismatch}. Figure \ref{NCC_MNIST_FMNIST_CIFAR} (Dashed Line) shows the \textbf{train} and \textbf{test} NCC mismatch of the network using the SVSL with the proposed hyperparameters, for MNIST, FashionMNIST, and CIFAR10. The visualization for the remaining Image datasets is given in the appendix. It is clear that for all datasets, and almost all layers, the NCC mismatch improves when using the SVSL. The same conclusions can be derived for all NLP datasets in Figure \ref{RTE_MRPC}.

We shall show the validity of Conjecture \ref{VSL_test}. Table \ref{tab:test_vsl} compares the test-performance of the vanilla Cross-Entropy (CE) loss with that of the SVSL on all datasets. This comparison is done at the IT, EOT, and also at the best Test-Epoch. We see that SVSL outperforms the vanilla CE at almost all stages of training. We also see that most datasets reach their best Test-Scores during TPT. Even when using the best possible Test Epoch, the SVSL loss achieves better or as-good results in all but one dataset. When the best Test Epoch is not achieved in the TPT, the scores achieved at the best Test-Epoch are comparable to the ones achieved at EOT. Practitioners in the field often look at regions of near-zero training-error, and use a validation set to choose the proper early-stopping criterion. This stage is formally given as the TPT, and hence a convincing method is to look at the performance mainly in this region. In these tasks we use the testing set as a proxy for the validation set. We see that even when allowing ourselves to look at all Test-Scores, the SVSL still achieves better performance on an array of tasks. All training graphs with both losses are given for the Vision datasets in Figures \ref{STL10_training_parts} and in the appendix.  In practice, one may use a hold-out validation/cross-validation set to choose the best epoch and achieve similar results to the maximal points in the plots.


\section{Conclusion}
In this paper, we expand the notion of NCC-Mismatch as proposed in \cite{Papyan2020PrevalenceON}. We describe how looking at intermediate layers of the network can assist in understanding the geometric phenomena that is Neural Collapse. This paper further expands these notions to NLP tasks, and shows common structure in the different modalities. We also show how encouraging inner-layer class-center consistency can assist in the training and generalization. We hope further research using these methods can continue to enrich the study in deepnets and their training paradigms. We further our discussion into possible usecases of results brought forward in the appendix.

\section*{Acknowledgements}
We would like to thank Gilad Fuchs for illuminating discussions during the preparation of this manuscript.

\bibliography{tmlr}
\bibliographystyle{icml2022}

\newpage
\appendix
\onecolumn

\section{Discussion and possible applications}
We point out two additional perspectives when looking at the results given in this work:
\begin{enumerate}
\item [(i)] \label{dataset_difficulty}\textbf{First-Layer NCC mismatch may suggest at the dataset ``dificulty'':} When looking at NCC mismatch in early layers of the network, there is an interesting thought experiment that can be suggested. On one end, if the beginning layers have a very low NCC mismatch during the TPT - this means that the network is already achieving very good class clustering early on. The earlier this happens, the less capacity the model has to achieve this clustering. When looking at Figure \ref{NCC_MNIST_FMNIST_CIFAR} (MNIST) - we see that in the first layer, we already have a mismatch of $\sim0.16$. Since MNIST is a very simple task, this might be intuitive. However, when looking at Figure \ref{NCC_MNIST_FMNIST_CIFAR} (CIFAR10) - we see that the first layer only reaches $0.6$ mismatch. This again, is intuitive as CIFAR10 is a hard task and we would not expect a few layers to be enough to properly cluster the features. This can be shown in several of the graphs along this paper. Perhaps this notion of thinking can aid in defining a concept of ``dataset dificulty'' for a certain model architecture. 

\item [(ii)]\textbf{NCC-Collapse may be useful for efficient inference in large models:} In most experiments shown in the paper, the NCC-Collapse does not happen solely in the penultimate model layer. In fact, in some architectures the collapse propagates a few layers back in the network. Suppose that for a trained network, the collapse occurs from all layers after layer $j$. This means that in order to get the prediction of the model on a new sample, we only need run a forward pass up to the $j$-th layer, and find the nearest train class-means (which needs to be computed once). In very deepnets, this can result in more efficient inference time.
\end{enumerate}
We hope that these points and others shown in this paper can encourage researchers to explore further the geometrical phenomena in intermediate layers.

\section{Hyper Parameters}\label{HP_params}
In Table ~\ref{tab:NLP_VSL_HP} we list the optimal Hyper-Parameters used in the experiments throught the paper.
\begin{table}\centering
\caption{SVSL-Hyperparametrs used for \textbf{Image} and \textbf{NLP sequence-classification} tasks, optimized using BoTorch \cite{balandat2020botorch}.}\vspace{0.1in}
\begin{tabular}{l|ll}
                                   & \textbf{$\alpha$} & \textbf{$\gamma$} \\ \hline
\textbf{MNIST}    & $1e-5$                                      & \multicolumn{1}{l}{Layer 1}              \\
\textbf{F-MNIST}  & $1e-5$                                      & \multicolumn{1}{l}{Layer 1}              \\
\textbf{STL10}    & $1e-5$                                      & \multicolumn{1}{l}{Layer 1}              \\
\textbf{CIFAR10}  & $4e-6$                                      & \multicolumn{1}{l}{AvgPool}              \\
\textbf{CIFAR100} & $1e-5$                                      & \multicolumn{1}{l}{Layer 3}              \\ \hline
\textbf{CoLA}     & $5e-6$                                      & \multicolumn{1}{l}{1}                    \\
\textbf{RTE}      & $1e-7$                                      & \multicolumn{1}{l}{11}                   \\
\textbf{MRPC}     & $1e-7$                                      & \multicolumn{1}{l}{11}                   \\
\textbf{SST-2}    & $4.132e-5$                                  & \multicolumn{1}{l}{10}                   
\end{tabular}
\label{tab:NLP_VSL_HP}
\end{table}

\section{Additional Experiments}
\begin{figure*}[t]
\begin{center}
   \includegraphics[width=0.95\linewidth]{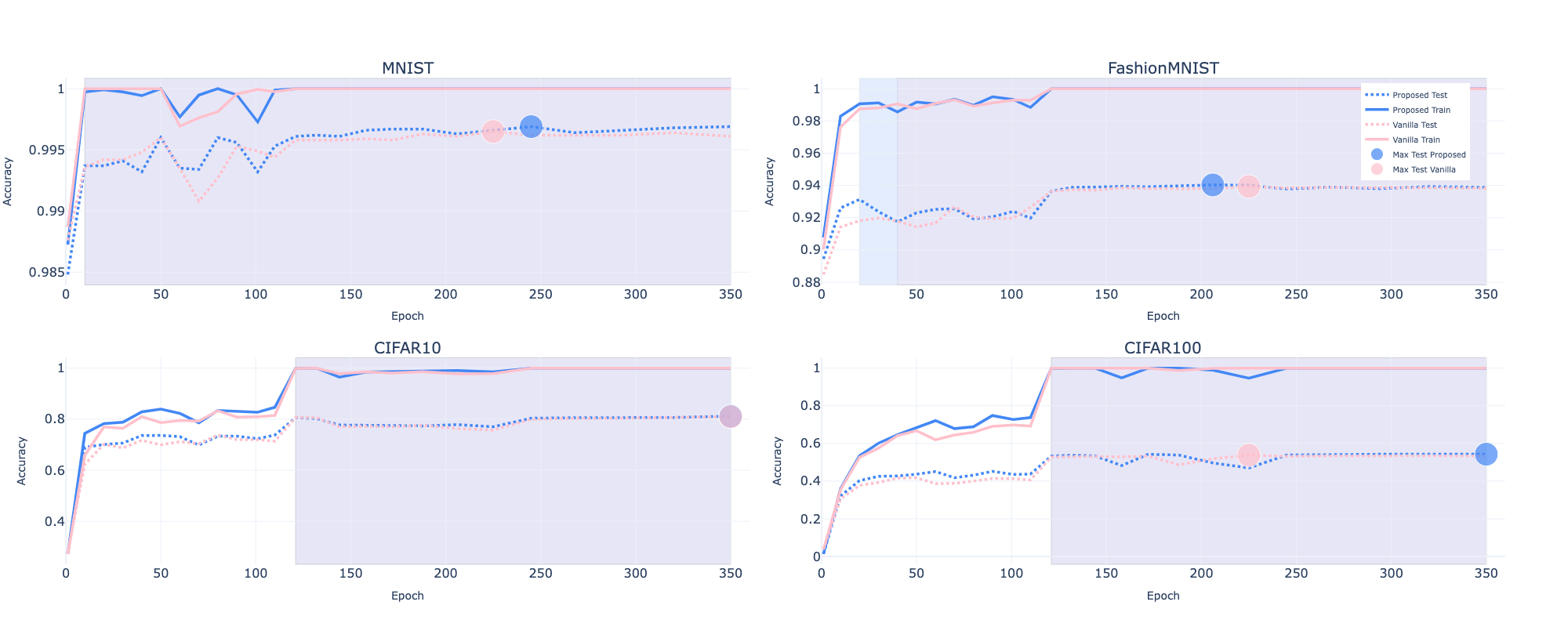}
\end{center}
   \caption{Optimization Procedures for Vision experiments: \textbf{MNIST}, \textbf{Fashion-MNIST}, \textbf{CIFAR10}, \textbf{CIFAR100}. The SVSL achieves higher performance(test and train) at most epochs. The TPT is marked with a shaded background, with colors according to the loss. For the Vision datasets, all models achieve best test performance during the TPT.}
\label{VISION_STAGES}
\label{fig:long}
\label{fig:onecol}
\end{figure*}
\begin{figure*}[t]
\begin{center}
   \includegraphics[width=0.95\linewidth]{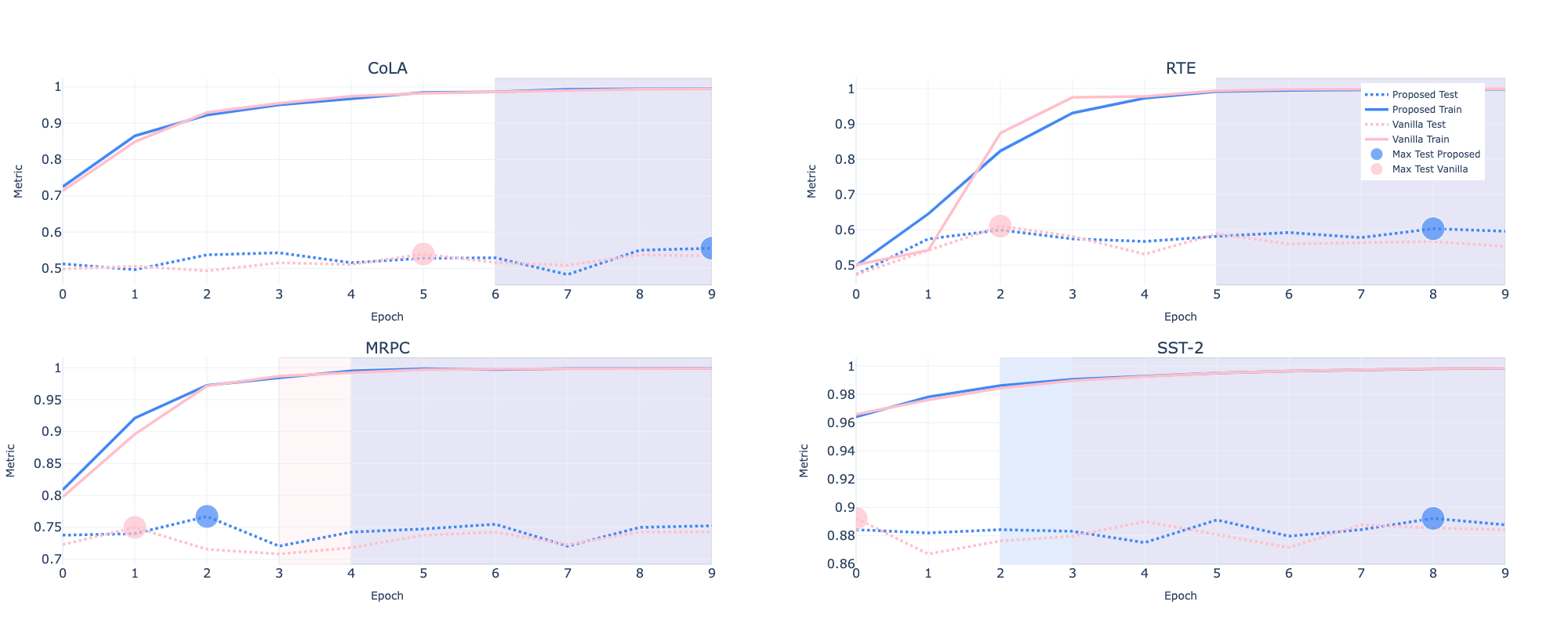}
\end{center}
    \caption{Optimization Procedures for Sequence-Classification experiments: \textbf{CoLA}, \textbf{RTE}, \textbf{MRPC}, \textbf{SST-2}. The TPT is marked with a shaded background, with colors according to the loss. The SVSL achieves higher performance(test and train) at most epochs.}
\label{NLP_STAGES}
\label{fig:long}
\label{fig:onecol}
\end{figure*}

\begin{figure*}[t]
\begin{center}
   \includegraphics[width=1.0\linewidth]{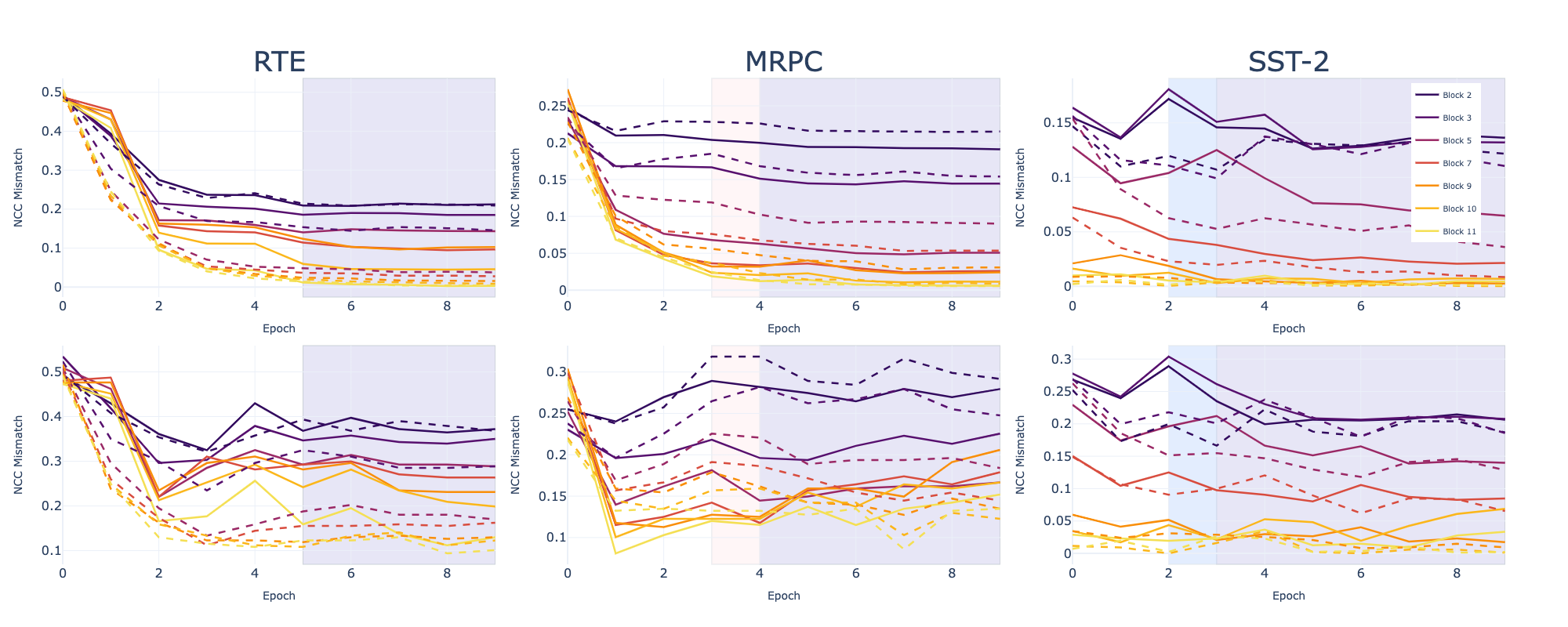}
\end{center}
   \caption{NCC mismatch for Sequence-Classification datasets: \textbf{RTE}, \textbf{MRPC} and \textbf{CoLA}, using both vanilla (\textbf{solid}) and SVSL(\textbf{dashed}) losses. \textbf{Top:} train NCC mismatch, \textbf{Bottom:} test NCC mismatch. We show only a subset of the transformer blocks for clearness. The shaded pink background shows the TPT for the vanilla loss experiment, and the blue for the SVSL. The background is shaded purple at epochs when both experiments are in the TPT.}
\label{RTE_MRPC}
\label{fig:long}
\label{fig:onecol}
\end{figure*}

\begin{figure}[t]
\begin{center}
   \includegraphics[width=1.0\linewidth]{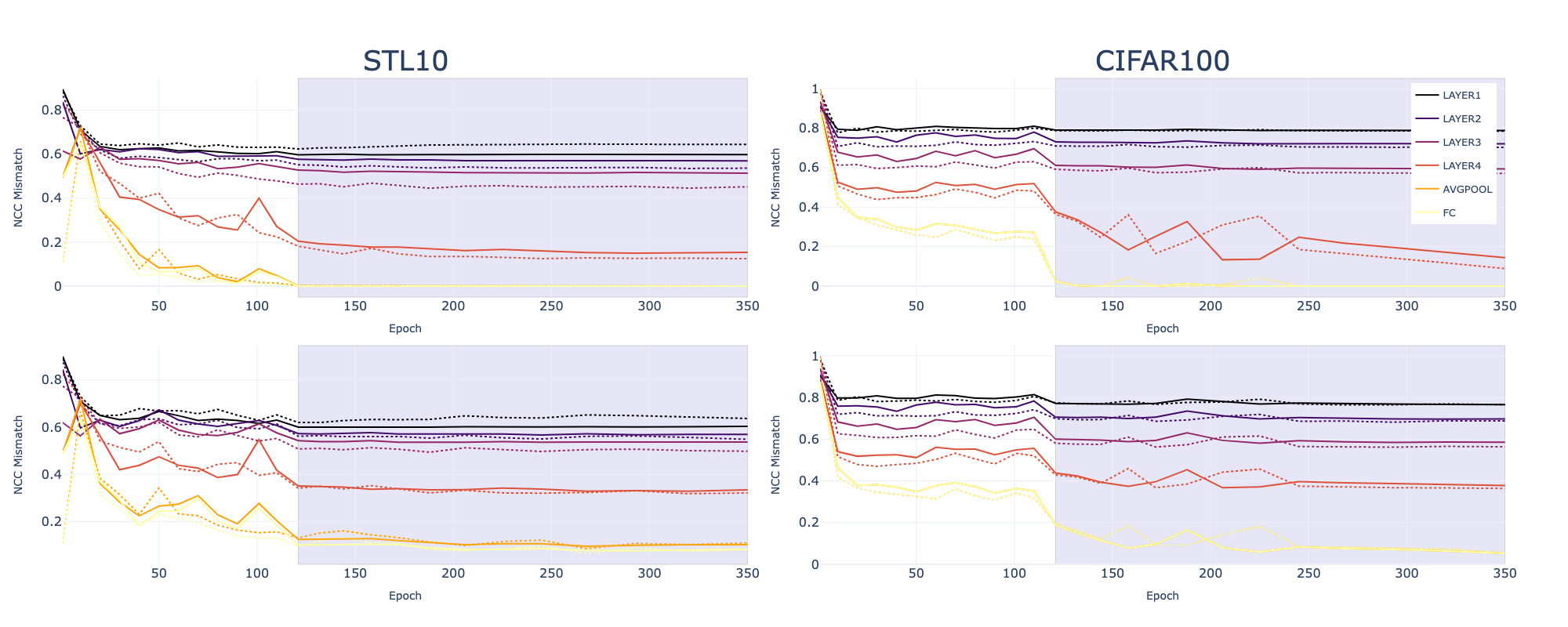}
\end{center}
   \caption{NCC mismatch for Vision Classification Datasets: \textbf{STL10} and \textbf{CIFAR100}, using both vanilla (\textbf{Solid}) and SVSL (\textbf{dashed}) losses. \textbf{Top:} train NCC mismatch, \textbf{Bottom:} test NCC mismatch. The IT for each of the losses is shown using a vertical dash-dot line.}
\label{STL10_CIFAR100}
\label{fig:long}
\label{fig:onecol}
\end{figure}

\begin{figure}[t]
\begin{center}
   \includegraphics[width=0.95\linewidth]{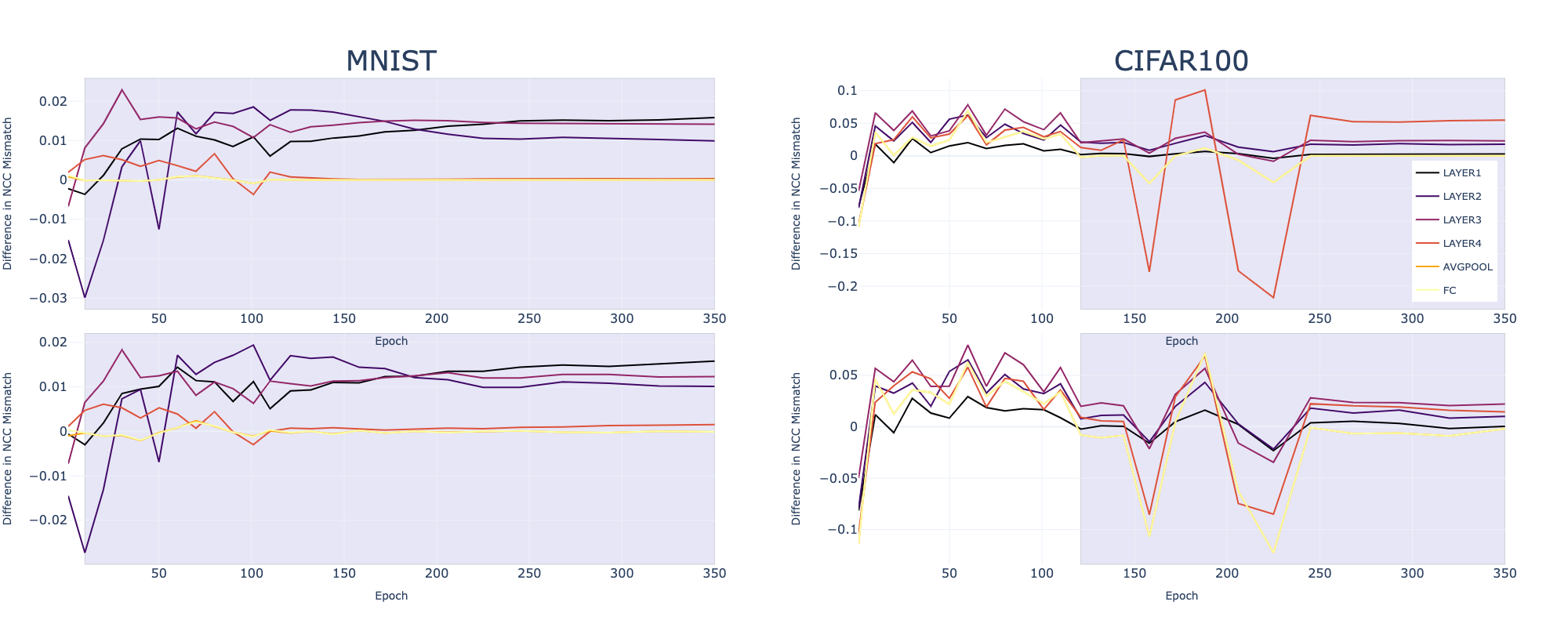}
\end{center}
   \caption{Intermediate layer NCC-Mismatch difference between the vanilla and SVSL losses for datasets: \textbf{MNIST} and \textbf{CIFAR100} in Vision. The $y$ axis shows $\Lambda_{\mathrm{Train, Vanilla}}^{\left(j\right)} - \Lambda_{\mathrm{Train, SVSL}}^{\left(j\right)}$ and $\Lambda_{\mathrm{Test, Vanilla}}^{\left(j\right)} - \Lambda_{\mathrm{Test, SVSL}}^{\left(j\right)}$ in the Top and Bottom resp. The shaded background represents the TPT in both losses as done in the remainder of the plots.}
\label{Vision_diff}
\label{fig:long}
\label{fig:onecol}
\end{figure}

\begin{figure}[t]
\begin{center}
   \includegraphics[width=0.95\linewidth]{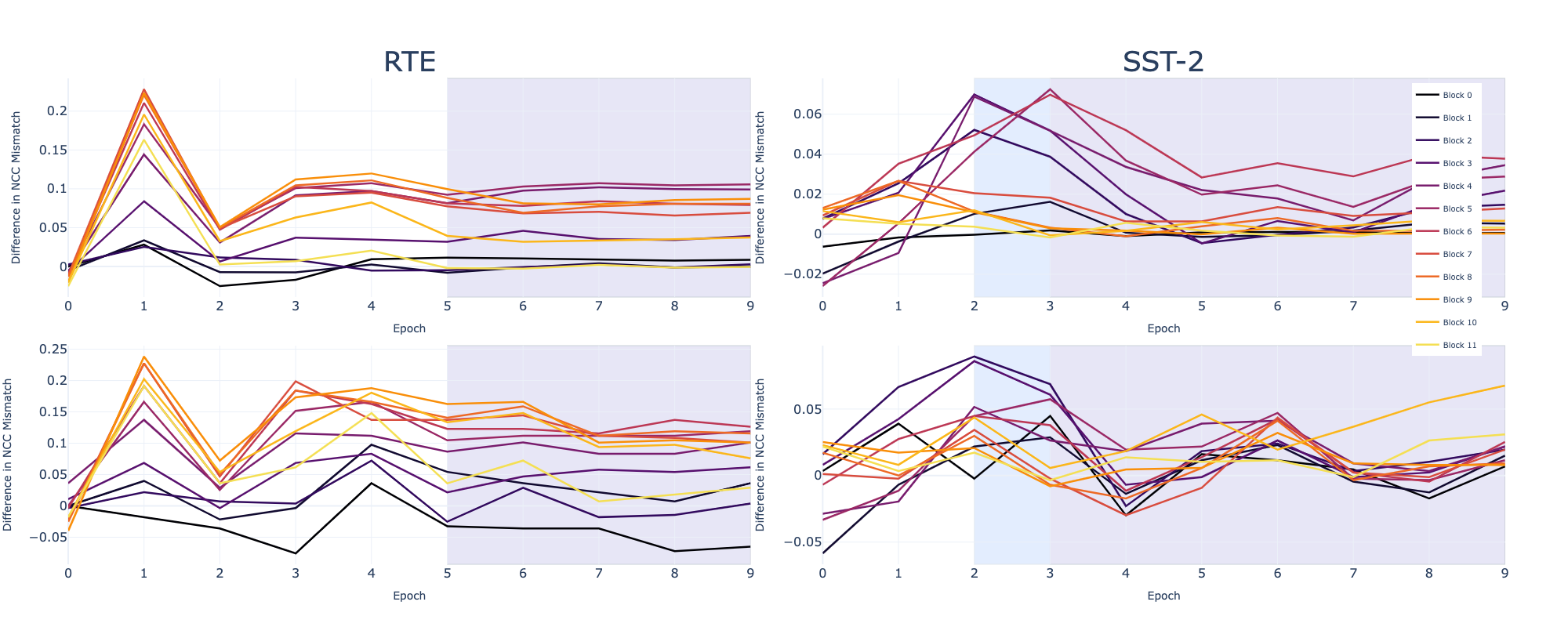}
\end{center}
   \caption{Intermediate layer NCC-Mismatch difference between the vanilla and SVSL losses for datasets: \textbf{RTE} and \textbf{SST-2} in Sequence-Classification. The $y$ axis shows $\Lambda_{\mathrm{Train, Vanilla}}^{\left(j\right)} - \Lambda_{\mathrm{Train, SVSL}}^{\left(j\right)}$ and $\Lambda_{\mathrm{Test, Vanilla}}^{\left(j\right)} - \Lambda_{\mathrm{Test, SVSL}}^{\left(j\right)}$ in the Top and Bottom resp. The shaded background represents the TPT in both losses as done in the remainder of the plots. }
\label{NLP_diff}
\label{fig:long}
\label{fig:onecol}
\end{figure}

Figures \ref{VISION_STAGES} and \ref{NLP_STAGES} show the training process and behavior in all Vision and Text datasets using both the Cross-Entropy and the SVSL loss functions. Figure \ref{RTE_MRPC} shows the NCC train and test metrics in the intermediate layers on the sequence classification datasets. In Figure \ref{STL10_CIFAR100} we show the NCC mismatch for the STL10 and the CIFAR100 Vision datasets. We see that these plots also show the characteristics shown in other datasets. We also see that for these datasets, the NCC mismatch in early layers of the network are higher. This matches the Discussion-Point \ref{dataset_difficulty}, as these datasets are considered more difficult than MNIST, Fashion-MNIST, and CIFAR10. Table \ref{tab:vision_changes_vanilla_intermediate} explicitly shows the values for Conjectures \ref{NCC_ordering} and \ref{NCC_improves_in_TPT} in the Vision Experiments throughout the paper. 
\\
In Figures \ref{Vision_diff} and \ref{NLP_diff} we introduce plots further telling of Conjecture \ref{VSL_Mismatch}. We plot the value of $\Lambda_{\mathrm{Train, Vanilla}}^{\left(j\right)} - \Lambda_{\mathrm{Train, SVSL}}^{\left(j\right)}$ and
$\Lambda_{\mathrm{Test, Vanilla}}^{\left(j\right)} - \Lambda_{\mathrm{Test, SVSL}}^{\left(j\right)}$ in the top and bottom rows resp. for intermediate layers $j$. These differences are the improvement in NCC match using the SVSL loss vs. the vanilla Cross Entropy. The Conjecture holds when the difference is greater or equal to zero. We show the intermediate NCC differences for sample Vision and Sequence Classification tasks.

\begin{table*}\centering
\caption{Comparing the \textbf{test} NCC mismatch at IT vs. EOT for Image Classification Datasets using the vanilla model architecture (scores are in percents)}\vspace{0.1in}
\label{tab:vision_changes_vanilla_intermediate}
\begin{tabular}{l|ll|ll|ll|ll|ll|ll}
\textbf{}        & \multicolumn{2}{l|}{\textbf{Layer 1}} & \multicolumn{2}{l|}{\textbf{Layer 2}} & \multicolumn{2}{l|}{\textbf{Layer 3}} & \multicolumn{2}{l|}{\textbf{Layer 4}} & \multicolumn{2}{l|}{\textbf{Avg. Pooling}} & \multicolumn{2}{l}{\textbf{FC}} \\
\textbf{Dataset} & \textbf{IT}       & \textbf{EOT}    & \textbf{IT}       & \textbf{EOT}    & \textbf{IT}       & \textbf{EOT}    & \textbf{IT}       & \textbf{EOT}    & \textbf{IT}         & \textbf{EOT}       & \textbf{IT}   & \textbf{EOT}  \\ \hline
MNIST            & 16.47             & \textbf{15.57}    & 11.16             & \textbf{10.05}    & 5.97              & \textbf{3.97}     & 2.33              & \textbf{0.34}     & 0.22                & \textbf{0.01}       & 0.16          & \textbf{0.01}  \\
F-MNIST    & 26.7              & \textbf{26.44}    & \textbf{20.97}    & \textbf{21.2}              & \textbf{15.97}    & 16.62             & 9.52              & \textbf{9.01}     & 0.3                 & \textbf{0.11}        & 0.3           & \textbf{0.13}   \\
STL10            & \textbf{59.74}    & \textbf{59.84}             & 57.62             & \textbf{56.96}    & 52.72             & \textbf{51.36}    & 20.42             & \textbf{15.32}    & \textbf{0.1}        & 0.4                  & \textbf{0.0}  & \textbf{0.02}            \\
CIFAR10          & 54.09             & \textbf{52.88}    & 43.48             & \textbf{40.8}     & 27.19             & \textbf{25.59}    & \textbf{9.17}     & 10.22             & 2.61                & \textbf{0.37}        & 2.61          & \textbf{0.37}            \\
CIFAR100         & 77.19             & \textbf{76.62}    & 70.61             & \textbf{69.76}    & 60.02             & \textbf{58.56}    & 43.76             & \textbf{37.82}    & 18.76               & \textbf{5.23}        & 18.76         & \textbf{5.23}
\end{tabular}
\end{table*}

\clearpage

\end{document}